\documentclass[conference]{IEEEtran}
\IEEEoverridecommandlockouts
\usepackage{cite}
\usepackage{amsmath,amssymb,amsfonts}
\usepackage{algorithmic}
\usepackage{graphicx}
\usepackage{textcomp}
\usepackage{xcolor}
\def\BibTeX{{\rm B\kern-.05em{\sc i\kern-.025em b}\kern-.08em
    T\kern-.1667em\lower.7ex\hbox{E}\kern-.125emX}}

\usepackage{epsfig} 
\usepackage{pgfplots} 
\usepackage{mathtools}
\usepackage{subcaption}
\usepackage{tikz}
\usepackage{enumitem}
\usepackage{booktabs}
\usepackage{array}
\usepackage{xspace}

\newcommand{\E}{\ensuremath{\operatorname{E}}\xspace}

\begin{document}

\title{Adversarial Latent Autoencoder with Self-Attention for Structural Image Synthesis
}
\author{\IEEEauthorblockN{1\textsuperscript{st} Jiajie Fan}
\IEEEauthorblockA{\textit{BMW Group} \\
Munich, Germany \\
jiajie.fan@bmw.de}
\and
\IEEEauthorblockN{2\textsuperscript{nd} Laure Vuaille}
\IEEEauthorblockA{
\textit{Technical University of Munich}\\
Munich, Germany \\
laure.vuaille@tum.de}\\
\and
\IEEEauthorblockN{3\textsuperscript{rd} Thomas B\"ack}
\IEEEauthorblockA{\textit{LIACS} \\
\textit{Leiden University}\\
Leiden, The Netherlands \\
t.h.w.baeck@liacs.leidenuniv.nl}
\and
\IEEEauthorblockN{4\textsuperscript{th} Hao Wang}
\IEEEauthorblockA{\textit{LIACS} \\
\textit{Leiden University}\\
Leiden, The Netherlands \\
h.wang@liacs.leidenuniv.nl}
}
\newcommand{\eg}{{ e.g.}}
\newcommand{\etal}{{ et~al.}}
\newcommand{\ie}{{i.e.}}
\newcommand{\etc}{{ etc.}}
\newcommand{\omegah}{{$\hat{\omega}$}}
\newcommand{\omegab}{{ $\bar{\omega}$}}

\maketitle



\begin{abstract}
Deep Generative Models (DGMs) have been successfully employed to synthesize general images, e.g., animals, human faces, and landscapes. This promising advancement leads to the idea of utilizing DGMs to generate novel structural designs, thereby facilitating industrial engineering processes. However, industrial design data, e.g., blueprints or engineering drawings, is fundamentally different from the images of natural scenes. They contain rich structural patterns and long-range dependencies, which are challenging for convolution-based DGMs to generate. We tackle this challenge by proposing the Self-Attention Adversarial Latent Autoencoder (SA-ALAE), which allows for generating realistic structure designs of complex engineering parts. With SA-ALAE, users can explore novel variants of an existing design and control the generation process by operating in the learned latent space. We showcase the potential of SA-ALAE by generating engineering blueprints in a real automotive design task.
\end{abstract}


\section{Introduction}
\label{introduction}
Generative Engineering Design (GED)~\cite{b1, b2} is a novel trend of generative design for a more function-oriented design exploration using existing designs and algorithms. GED derives novel designs depending on factors such as the initial structure, desired performance, and specific project requirements. As an automated exploration of industrial designs, GED shows great potential to assist engineers in developing complex structures and speeding up industrial development processes. Deep Generative Models (DGMs)~\cite{b3, b4, b5} seem to be a natural choice to facilitate GED, given their compelling performance in synthesizing natural images, such as landscapes, human faces, and animals. However, the implementation of DGMs as such is unlikely to bring much benefit to industrial engineering processes. 

One reason is that most DGMs fail to provide a controllable generation process. Recent top-performing DGMs tend to use Generative Adversarial Networks (GANs)~\cite{b4, b5} as a framework, which derives a single-generator sampling procedure and typically generates a novel image from a random noisy input. Novel models~\cite{b8,b9,b10} have been proposed to condition GANs on additional input variables, hereby enabling fine-grained control over the generation process. These input variables are predefined attributes, such as gender, hairstyle, or face shape, in the context of human face generation. However, it is difficult to preset input conditioning variables in industrial use cases because of the lack of universal conventions used to define design parameters and large, consistent labeled data. Furthermore, the relationship between the parameters and the engineering target often requires complex numerical methods to be resolved. It is thus difficult to define relevant class labels to condition the generation process. Alternatively, Adversarial Latent Autoencoder (ALAE)~\cite{b11} utilizes GAN's adversarial training strategy but provides an encoder-decoder network, which allows for a mapping between image space and latent space. Through our research, we have observed the huge potential of this structure in facilitating generative engineering design for industrial use cases, \eg, ALAE can extract latent variables from a known design and utilize them to derive new designs through controlled modifications.

\begin{figure}[t]
\centering
\includegraphics[width=0.48\textwidth, trim=2.5mm 0mm 3mm 5mm, clip]{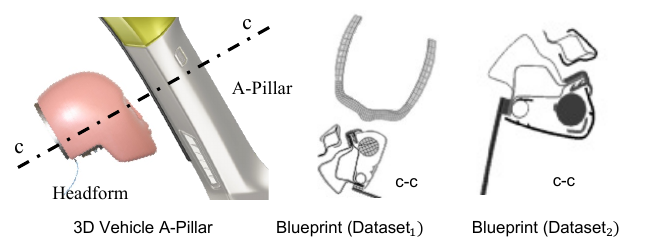}
\caption{Vehicle A-Pillar 3D model (\textit{left}) and various versions of A-Pillar blueprints obtained by extracting cross-sections from the 3D object.}
\label{fig:cross-section_example}
\end{figure}

Another challenge lies in the fundamental difference between structural images of engineering modeling and images of natural scenes. The latter typically consists of rich textures and continuously changing color gradients, while the former is defined by geometric patterns, such as distinct edges, sparse shapes, and their associated long-range dependencies. Most DGMs rely heavily on the convolutional operator, which excels in synthesizing realistic textures in local neighborhoods but fails to model large objects~\cite{b12, b13}. With convincing results, recent research~\cite{b13,b14} demonstrates the potential of a self-attention mechanism in enabling DGMs to capture and model large-scale features and long-range dependencies.

Finally, we propose a novel model called Self-Attention Adversarial Latent Autoencoder (SA-ALAE), which builds on the ALAE architecture and employs additional techniques, e.g., Spectral Normalization~\cite{b22} and Residual Network (ResNet)~\cite{b26}, to enhance the stability of the adversarial training process. Most importantly, SA-ALAE incorporates the attention mechanism to generate intricate structural features. We show the potential of SA-ALAE by applying it in the context of automotive design exploration, where structural design is represented in engineering blueprints. SA-ALAE is developed on the first version of the engineering blueprints, as shown in the middle in Fig.~\ref{fig:cross-section_example}. Since SA-ALAE yields convincing results, we retrain and test SA-ALAE on the second version of the blueprints, shown on the right in Fig.~\ref{fig:cross-section_example}, where SA-ALAE has also achieved impressive quality of generated designs. To measure the quality of the synthesized blueprints, we utilize Fréchet inception distance (FID) \cite{b15}, which quantifies the inception distance between the source dataset and the generated image set. We further show with one experiment that SA-ALAE enables a controlled generation process by semantically modifying the encoded latent variables. The result of this experiment showcases the potential of SA-ALAE in understanding the structural information in engineering blueprints and exploring novel design alternatives in desired ways.

\section{Related work}\label{sec:related-work}

Within the domain of Deep Generative Models (DGMs), Generative Adversarial Networks (GANs)~\cite{b4, b5} have achieved state-of-the-art performances in many image synthesis tasks~\cite{b6, b7}. Novel models, e.g., Conditional Generative Adversarial Networks (CGANs)~\cite{b8}, StyleGAN~\cite{b9} and InfoGAN~\cite{b10}.
Inspired by the use of latent space to control the generation process~\cite{b5, b16, b17}, progress has been made in enabling GANs to map images to latent variables. The most successful trend is to combine GANs with encoders, which has already yielded convincing models such as BiGAN~\cite{b18}, Adversarially Learned Inference (ALI)~\cite{b19} and Adversarial Latent Autoencoder (ALAE)~\cite{b11}. After identifying and understanding the latent variables after training, semantic modification in the latent space can influence the generation process in desired ways. Among BiGAN, ALI, and ALAE, the latter achieves the best image quality when tasked to reconstruct MNIST~\cite{b20} digits and also the highest accuracy when applying MNIST classification to its generated outputs~\cite{b11}. After training, ALAE provides an encoder-decoder inference, allowing ALAE to map images and the latent space. 

Moreover, GANs are known of suffering from training instability~\cite{b6, b21, b22}. Many methods focus on solving this issue, \eg, zero-centered gradient penalty~\cite{b23}, Spectral Normalization~\cite{b22}, Batch Normalization~\cite{b24, b25}, etc. Taking Residual Networks (ResNet)~\cite{b26} as a backbone enables the training of deep architectures without suffering from vanishing gradients and degradation in performance.

In Generative Engineering Design (GED)~\cite{b27, b28, b29, b30}, DGM has shown its potential in synthesizing airfoil shapes from the UIUC database~\cite{b31}, in which the shape is a single 2D curve. However, real-world engineering designs are often much more complex, containing multiple clear edges, sparse shapes, and long-range geometric constraints. Generating such complex design images is still challenging for convolution-only neural networks because the convolutional operator can only process information within a local neighbourhood~\cite{b12,b13}. Self-Attention Generative Adversarial Network (SAGAN)~\cite{b13} introduces the self-attention mechanism into GANs. The implemented mechanism, Scaled Dot-Product Attention~\cite{b14}, constructs an attention map by computing the attention weights between each feature vector and the other feature vectors. This attention map presents the relationships among input features. With this attention map, the model gains a global understanding of the input features and can efficiently capture long-range dependencies.


\section{Methods}\label{sec:methods}
In this work, we select Adversarial Latent Autoencoder (ALAE)~\cite{b11} as a baseline framework for its flexible sampling process. We enhance the model robustness and develop an improved version by introducing modern techniques, e.g., Spectral Normalization (SN)~\cite{b22}, ResNet~\cite{b26}, etc. We further implement the self-attention mechanism~\cite{b13, b14} to enable the generation of structural patterns.

\subsection{Background}
\label{sec:backgorund}
\begin{figure}[t]
   \fontsize{12}{12}\selectfont
     \begin{subfigure}[b]{0.48\textwidth}
         \centering
        \includegraphics[width = \textwidth]{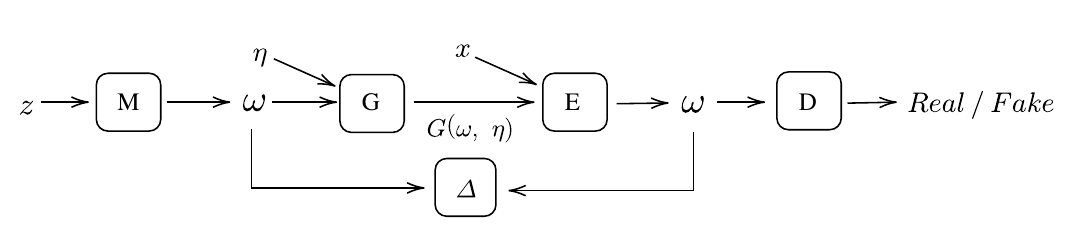}
         \caption{Training}
         \label{fig:Training}
     \hspace{0.2in}
     \end{subfigure}\\
     \begin{subfigure}[b]{0.48\textwidth}
         \centering
        \includegraphics[width =\textwidth]{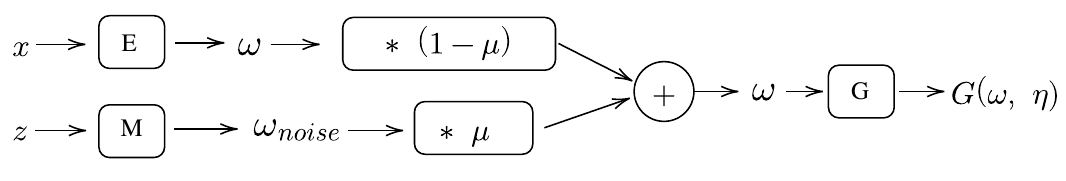}
         \caption{Sampling}
         \label{fig:sampling}
     \end{subfigure}\\
     \hfill
        \caption{SA-ALAE in directed graphs. ALAE and SA-ALAE share an identical training strategy. The sampling procedure of SA-ALAE utilizes the off-the-shelf mapper $M$ to sample additional noisy latent variables $\omega_{noise}$, hereby introducing randomness in the sampling.}
        \label{fig:ALAE_structure}
\end{figure}

The architecture of ALAE is shown in Fig.~\ref{fig:Training}. The ALAE implements a mapper $M$ that generates $d$-dimensional latent variables $\vec{\omega}\in\mathcal{W}\subseteq \mathbb{R}^d$ from standard multivariate Gaussian random noise $\vec{z}\sim\mathcal{N}(\vec{0}, \mathbf{I}), \vec{z}\in \mathbb{R}^k$. 
The latent variables $\vec{\omega}$ serves as input to both the discrimination map ($D\colon\mathcal{W}\rightarrow \mathbb{R}$) and the generation map ($G\colon\mathcal{W}\times \mathbb{R}^d \rightarrow\mathcal{X}$), where $\mathcal{X}$ stands for the space of data points. In addition, we denote by $\vec{x} \sim \mathcal{D}$ the distribution of the real data points. To generate a data point, the map $G(\vec{\omega}, \vec{\eta})$ takes as input a latent variable $\vec{\omega}$ and an optional Gaussian noise $\vec{\eta}\in\mathbb{R}^d$. Similar to an autoencoder, ALAE transforms a data point data $\vec{x}$ to its latent representation with an encoder map ($E\colon\mathcal{X}\rightarrow\mathcal{W}$), where $\vec{x}$ can be either a real data or generated from the map $G$. Afterward, the encoded data point is validated with the discriminator $D$.

Compared to GANs, the composition $G\circ M$ is equivalent to the generator of GAN, and $D\circ E$ serves as the discriminator.
In this view, one can implement an adversarial training of all the maps, \ie, to solve the following minimax problem w.r.t.~the weights of $M,G,E,D$:
\begin{equation} \label{eq:minimax}
    \min_{M,G}\max_{E,D}V(G\circ M, D\circ E),
\end{equation}
where the value function $V$ takes the following form: $V(G\circ M, D\circ E) = \E_{\vec{x}\sim \mathcal{D}} f(D\circ E(\vec{x})) + \E_{\vec{z}\sim \mathcal{N}(\vec{0}, \mathbf{I})} f(-D\circ E \circ G\circ M(\vec{z}))$. According to~\cite{b11}, ALAE implements $f(\cdot)$ as a \textit{SoftPlus} function~\cite{b32}, \ie, $f(t) = \textit{softplus}(t) = \log(1 + \exp(t))$.

Also, one can consider the maps $G\colon \vec{\omega} \mapsto \vec{x}$ and $E\colon \vec{x} \mapsto \vec{\omega}$ as the encoder and decoder of the latent variable $\vec{\omega}$, respectively. Ideally, we wish to have $E \circ G=\operatorname{Id}_{\mathcal{W}}$. Practically, we minimize the discrepancy between the probability distribution of $\vec{\omega}$: $p_{\vec{\omega}} = p_{\vec{z}}(M^{-1}(\vec{\omega}))$ and its pushforward under $E \circ G$, i.e., $(E \circ G)_*(p_{\vec{\omega}}) = p_{\vec{z}}(M^{-1} \circ G^{-1} \circ E^{-1} (\vec{\omega}))$:
\begin{equation} \label{eq:discrepancy}
    \min_{E,G}\Delta(p_{\vec{\omega}}, (E \circ G)_*(p_{\vec{\omega}})),
\end{equation}
where $\Delta$ is a reconstruction error, e.g., Mean Square Error (MSE) or KL-divergence.
In the training phase, the optimization task alternates between~\eqref{eq:minimax} and~\eqref{eq:discrepancy}.

After training, the inference/generation task requires the maps $E, G, M$, which take a source image $\vec{x}$ and a noise $\vec{\eta}\sim\mathcal{N}(\vec{0}, \mathbf{I})$ as inputs and generates new images. Additionally, feeding sampled noise $z$ to the pair of networks $(M, G)$ can generate random samples.

\subsection{Improved ALAE}
\label{sec:stabilizingMethods}
The vanilla ALAE faces a major issue in its training for the following reasons. First, the ALAE implements the adversarial training strategy in image space, which exposes it to the same training instability commonly encountered by GANs~\cite{b6,b23,b21}. We implement the following techniques to improve the model robustness and propose a novel version of the ALAE called ALAE$_{improved}$. Firstly, to ensure that generator $G$ receives effective gradients for the optimization, we need to prevent the computed gradients from vanishing or exploding. For this purpose, we apply Spectral Normalization (SN)~\cite{b22} in encoder $E$ and discriminator $D$. Spectral Normalization is a powerful method that constrains the Lipschitz continuity of the functions optimized by $E$ and $D$ through layer-wise control of the spectral norm, thereby constraining the scale of the gradients. Spectral Normalization replaces every given weight matrix $W$ by $W_{\text{sn}}$ with the formula:
\begin{equation}
W_{\text{sn}} = \frac{W}{\sigma(W)},
\end{equation}
\noindent
where $\sigma(W)$ is the greatest singular value of $W$. 
Moreover, to stabilize the adversarial training, we implement batch normalization in generator $G$, allowing for a larger range of learning rates and learning the data distribution more efficiently. 
Lastly, to address the degradation caused by increasing the depth of the neural network, we modify generator $G$ and encoder $E$ of ALAE to become Residual Networks (ResNet). ResNet features in the so-called residual blocks of convolutional operators contain a ``skip connection'' that can bypass the error information in back-propagation, hence alleviating the vanishing gradient problem in deeper architectures.

\subsection{Self-Attention Adversarial Latent Autoencoder (SA-ALAE)}
\label{sec:SAALAE}
Convolution-based DGMs generate high-quality real-world images. However, DGMs have difficulty in modeling structural objects in engineering design images. The main cause is possibly that the objects in structural images can be sparsely located but are still strongly geometrically connected, e.g., in bicycle design, both wheels should be connected by an axle and lie on the ground. In this simple example, the pixels describing each wheel depend on one another but are far away in the image. However, as a local operator on images, the convolution mechanism cannot capture long-range dependencies, suggesting that using convolutional layers alone is unsuitable for generating engineering design images. Self-Attention Generative Adversarial Network (SAGAN)~\cite{b13} leverages the self-attention mechanism to gain a global understanding of the input features, which enables the model to capture and model long-range dependencies efficiently. Following the success of SAGAN, we introduce the same mechanism respectively into generator $G$ and encoder $E$ in the ALAE$_{improved}$ framework and propose a new model, Self-Attention Adversarial Latent Autoencoder (SA-ALAE).

To achieve the stochastic generation of a novel image from a source image in the vanilla ALAE, the latent variables $\omega$ encoded from the original design are perturbed with sampled noise $\vec{\eta}\sim\mathcal{N}(\vec{0}, \mathbf{I})$. We note that this approach compromises the quality of the encoded latent variables, which generates blurry images. To avoid this, SA-ALAE leverages the trained map $M$ to sample random latent variables as a noisy perturbation. The novel sampling procedure of SA-ALAE is displayed in Fig.~\ref{fig:sampling}. Adding randomly sampled latent variables as an independent noise protects the quality of perturbed latent variables so that the stochastic encoder-decoder inference does not harm the quality of the generated image. 
The calculation of the latent variables, used to explore the variants of existing designs, is based on the following formula:
\begin{equation}
 \label{weightedMeanFormulate}
\vec{\omega} = (1-\mu)E(\vec{x})  + \mu M(\Vec{z}),
\end{equation}
\noindent
where $\mu \in [0,1]$ is a tunable parameter, $\vec{x}$ refers to an original design and $\vec{z}\sim\mathcal{N}(\vec{0}, \mathbf{I})$ is sampled noise.

\begin{figure}[t]
     \centering
    \includegraphics[width=0.45\textwidth,trim=1mm 2mm 0mm 4mm, clip]{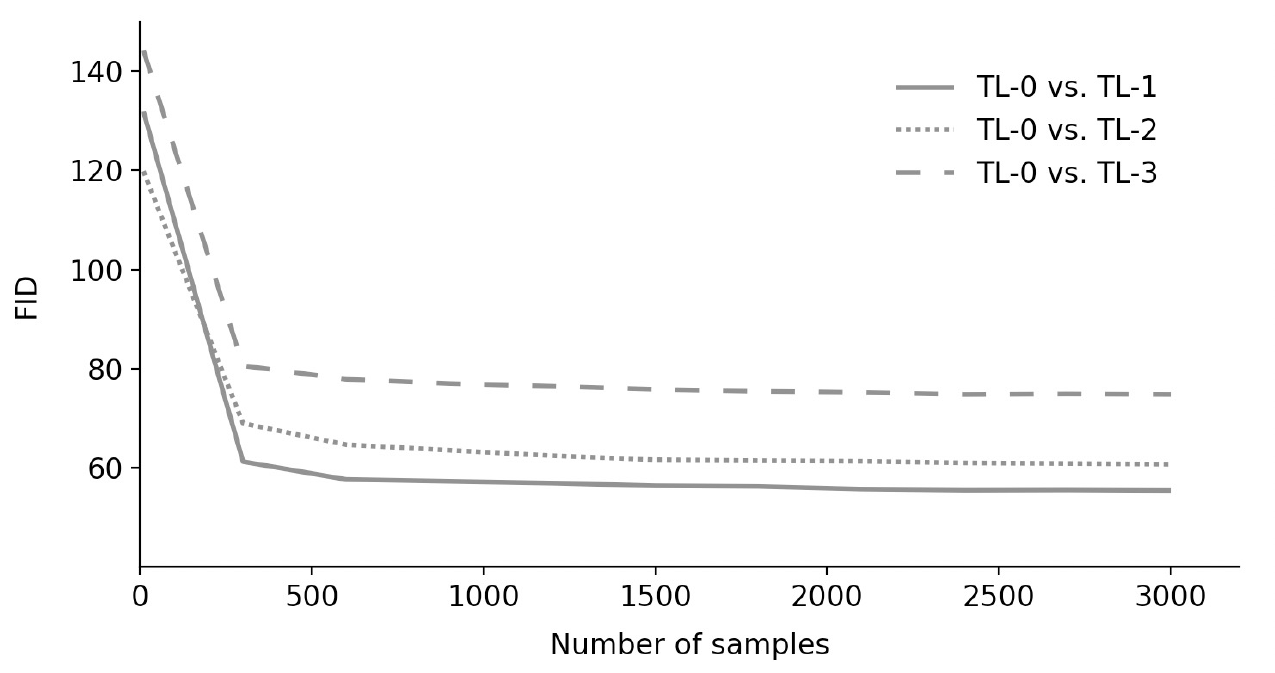}
    \caption{FID measured in terms of sample size. TL-0,1,2, and 3 stand for various target locations on vehicle A-Pillars where the cross-sections are taken. For this experiment, all blueprints are from Dataset$_1$.}
    \label{fig:NumberOfSamples}
\end{figure}

\section{Experiments}

\subsection{Datasets}
\label{sec:datasets}

To evaluate the performance of SA-ALAE for industrial design exploration, we trained the model to generate blueprints of vehicle parts from a real industrial design task. The whole dataset consists of $157\,234$ blueprints from various automotive parts, e.g., A-Pillars, B-Pillars, upper roofs, and side rails, etc. The blueprints are grayscale pixel-based images with a resolution of $256\times256$. They are produced by extracting cross-sections from the vehicle structures at various target locations. Among them, the sub-dataset that contains blueprints from A-Pillars is the largest set with 12\,876 samples, which is used in this work as a simpler dataset due to its lower diversity. 

In our work, we consider two types of blueprints: Dataset$_1$ contains some irrelevant information/objects, such as a head-form model used in a specific engineering process and wireframes, as shown in the middle in Fig.~\ref{fig:cross-section_example}; Dataset$_2$ is a preprocessed dataset in which the head form has been removed, images are centered on the structure, and specific areas are filled with continuous grayscale color, as shown on the right in Fig.~\ref{fig:cross-section_example}. Updating the Dataset$_1$ to Dataset$_2$ is helpful with evaluating the peak performance of the generative model in generating detailed structures and testing the model's applicability on various datasets.

\subsection{Evaluation methods}
\label{sec:EvaluationMethods}

In the area of DGMs, developers often conduct qualitative analysis because there is no ground truth. In our work, we compare various models by assessing the visual quality of the generated images in terms of blurriness, detailed structure, and readability. Additionally, we introduce a quantitative assessment to evaluate model performance, where we follow the recent research in the domain of image generation~\cite{b3,b9, b11,b13} and utilize Fréchet Inception Distance (FID)~\cite{b15} to quantify the improvement among models. 

As a rule of thumb, FID is calculated with $50k$ generated samples to evaluate the performance of a trained model~\cite{b15}. However, generating images of this high number incurs significant computational costs. As shown in Fig.~\ref{fig:NumberOfSamples}, the measured FID values are very sensitive to the sample size until 1\,000, where we observe a plateau. Based on this observation, taking a smaller set of 1\,000 images for empirical assessment of our dataset is safe. In quantitatively evaluating a target model, the model samples 1\,000 images for random noisy inputs, and the FID is measured between the sampled images and 1\,000 corresponding source images.

Given that our work is conducted on a domain-specific dataset without any previous comparative studies, it is essential to establish an FID baseline for evaluation. In our work, we measure the FID between two sets of real A-Pillar blueprints as the FID baseline. Specifically, we compare 1\,000 cross-sections taken from two neighboring target locations. Since the structural difference is minimal between two neighboring target locations, the measured FID values serve as good baselines for the evaluation, where we obtain an FID of $55.40$ for Dataset$_1$ and an FID of $35.58$ for Dataset$_2$.



\begin{table}[t]
\fontsize{8}{10}\selectfont
\caption{FID scores. The lower the better.}
\label{tab:FIDResults}
\begin{center}
\begin{tabular}{p{6.5cm} m{1cm}}
\toprule
\textbf{Objective} & \textbf{FID}$\downarrow$ \\ \hline
\textbf{Using Dataset$_1$ as source}  \\\hline
Baseline & $55.40$\\ 
ALAE~\cite{b11} & $284.01$ \\ 
ALAE$_{improved}$ & $135.52$ \\ 
SA-ALAE [Ours]& $90.09$  \\ 
SA-ALAE$_{sub}$ [Ours]& $\textbf{48.78}$  \\ \hline
\textbf{Using Dataset$_2$ as source}  \\\hline
Baseline & $35.58$\\ 
SA-ALAE$_{sub}$ [Ours]& $\textbf{30.10}$  \\ 
\bottomrule
\end{tabular}
\footnotetext{Bold numbers indicate the best results in this experiment}
\end{center}
\end{table}


\subsection{Training configurations}
\label{sec:configurations}
For the training, the initial learning rates for the optimizations $lr_{(D,E)}$, $lr_{(M,G)}$, $lr_{(G,E)}$ are $10^{-4}$, $2\times 10^{-4}$, and $10^{-4}$, respectively. The batch size is set to $128$; the dimension of the Gaussian random noise $\vec{z}$ is set to 128; the dimension of the latent variable $\vec{\omega}$ is $64$. We apply a self-attention layer after each upsampling and downsampling module, where the size of the output features is $16\times16$. From each target dataset used, we randomly select 1\,000 images as test data and 100 images as validation data, while the remaining images serve as training data. For computational efficiency, the FID is measured between 100 source images and 100 generated images after every epoch during the training to evaluate the model performance. The training procedure is terminated if the measured FID does not improve within 20 epochs. 

\subsection{Results}
In this work, we compare various models for Dataset$_1$:
\begin{itemize}
    \item ALAE: training vanilla ALAE~\cite{b11} model with blueprints of all parts from scratch;
    \item ALAE$_{improved}$: training improved version of ALAE model with blueprints of all parts from scratch;
    \item SA-ALAE: training our SA-ALAE model with blueprints of all parts from scratch;
    \item SA-ALAE$_{sub}$: fine-tuning the trained SA-ALAE model, that is trained on blueprints of all parts, with blueprints of A-Pillar;
\end{itemize}
As for Dataset$_2$, instead of using blueprints from all parts, we train the SA-ALAE model only on blueprints of A-Pillars from scratch, hereby producing the model SA-ALAE$_{sub}$. The measured FIDs are listed in Table~\ref{tab:FIDResults} and compared with the FID baselines determined in section~\ref{sec:EvaluationMethods}. A lower FID demonstrates a higher similarity to the source dataset.

\begin{figure}[t]
     \centering
          \begin{subfigure}[b]{0.48\textwidth}
        \includegraphics[width = 0.98\textwidth,trim=5mm 0mm 0mm 1mm, clip]{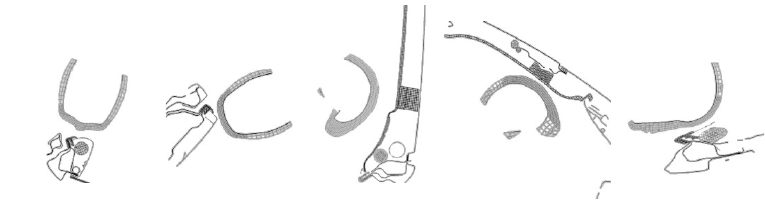}
         \caption{\textbf{Source blueprints (of all parts) from Dataset$_1$}}
         \label{fig:InputImagesWhole}
     \end{subfigure}\\
     \hfill
     \begin{subfigure}[b]{0.48\textwidth}
        \includegraphics[width = 0.98\textwidth,trim=5mm 0mm 0mm 1mm, clip]{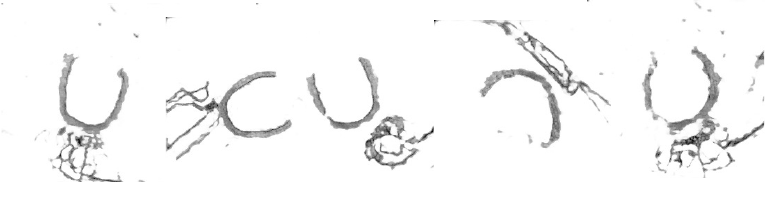}
         \caption{\textbf{Blueprints reconstructed by vanilla ALAE~\cite{b11}}}
     \end{subfigure}\\
     \hfill
     \begin{subfigure}[b]{0.48\textwidth}
        \includegraphics[width = 0.98\textwidth,trim=5mm 0mm 0mm 1mm, clip]{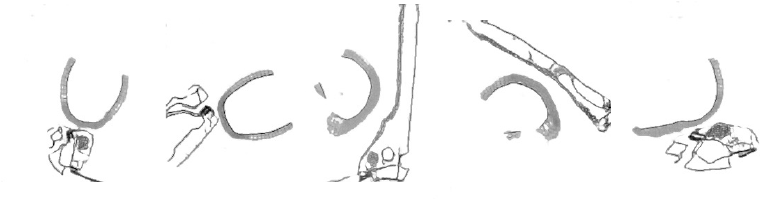}
         \caption{\textbf{Blueprints reconstructed by SA-ALAE [Ours]}}
     \end{subfigure}\\
     \hfill
    \caption{Examples of source blueprints from Dataset$_1$ and generated blueprints.}
    \label{fig:figure_generated_images_D1}
\end{figure}

\begin{figure}[h]
     \centering
          \begin{subfigure}[b]{0.48\textwidth}
        \includegraphics[width = 0.98\textwidth,trim=1mm 0mm 1mm 5mm, clip]{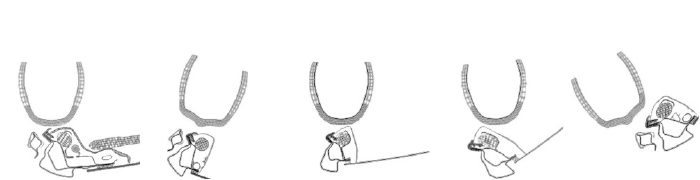}
         \caption{\textbf{Source blueprints (A-Pillar) from Dataset$_1$}}
         \label{fig:InputImagesAP}
     \end{subfigure}\\
     \hfill
          \begin{subfigure}[b]{0.48\textwidth}
        \includegraphics[width =0.98\textwidth,trim=1mm 0mm 1mm 5mm, clip]{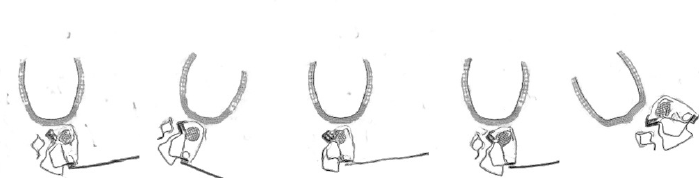}
         \caption{\textbf{Blueprints reconstructed by SA-ALAE}$_{sub}$}
         \label{fig:SA-ALAEaph}
     \end{subfigure}\\
     \hfill
        \caption{Examples of source blueprints of A-Pillar from Dataset$_1$ and blueprints generated by SA-ALAE$_{sub}$.}
        \label{fig:figure_generated_images_D1_AP}
\end{figure}

\begin{figure}[t]
    \fontsize{8}{10}\selectfont
     \begin{subfigure}[b]{0.48\textwidth}
         \centering
        \includegraphics[width = 0.98\textwidth,trim=3mm 6mm 3mm 1mm, clip]{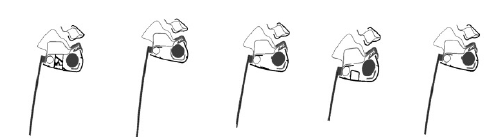}
         \caption{\textbf{Source blueprints (A-Pillar) from Dataset$_2$}}
         \label{fig:InputImagesV2}
     \end{subfigure}\\
     \hfill
          \begin{subfigure}[b]{0.48\textwidth}
         \centering
        \includegraphics[width =0.98\textwidth,trim=3mm 6mm 3mm 1mm, clip]{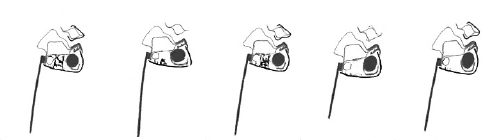}
         \caption{\textbf{Blueprints reconstructed by SA-ALAE}$_{sub}$}
         \label{fig:SA-ALAEIaph}
     \end{subfigure}\\
      \hfill
          \begin{subfigure}[b]{0.48\textwidth}
         \centering
        \includegraphics[width =0.98\textwidth,trim=3mm 12mm 3mm 1mm, clip]{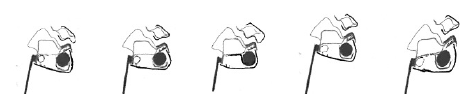}
         \caption{\textbf{Blueprints randomly generated by SA-ALAE}$_{sub}$}
         \label{fig:SA-ALAEIaph}
     \end{subfigure}\\
     \hfill
        \caption{Examples of source blueprints of A-Pillar from Dataset$_1$ and blueprints generated by SA-ALAE$_{sub}$.}
        \label{fig:figure_generated_images_new}
\end{figure}

Trained on the whole Dataset$_1$, ALAE$_{improved}$ significantly reduces the FID value by $52\%$ compared to the vanilla ALAE model, while SA-ALAE shows a further improvement over ALAE$_{improved}$, with a $34\%$ decrease in FID. As generated images shown in Fig.~\ref{fig:figure_generated_images_D1}, SA-ALAE can accurately reconstruct the input structural design, whereas vanilla ALAE generates blueprints with blurry textures. However, the designs generated by SA-ALAE are insufficient to support the engineering process due to the presence of unrecognizable details. We conjecture that it might be due to the diversity of the source vehicle parts represented in the blueprints. Given that the A-pillar has the largest amount of blueprints and SA-ALAE performs best on the A-pillar blueprints in the previous results, we retrain the SA-ALAE model only on the engineering blueprints from vehicle A-Pillars. As a result, SA-ALAE$_{sub}$ achieves an FID value of $48.78$, which is $12\%$ superior to the baseline. As blueprints generated by SA-ALAE$_{sub}$ displayed in Fig.~\ref{fig:figure_generated_images_D1_AP}, SA-ALAE$_{sub}$ can generate detailed blueprints and successfully captures and represents structural information, e.g., rotations, distances, and shapes. When using blueprints of A-Pillar from Dataset$_2$ as training data, our SA-ALAE$_{sub}$ outperforms the baseline by $15\%$ in terms of FID. As shown in Fig.~\ref{fig:figure_generated_images_new}, SA-ALAE$_{sub}$ generates high-quality blueprints corresponding to the input designs and containing recognizable structural details. In addition to the reconstruction function, we evaluate the ability of SA-ALAE$_{sub}$ in randomly generating A-Pillar designs, which yields a sufficient diversity of novel designs with detailed structures displayed in Fig.~\ref{fig:figure_generated_images_new}.


\begin{figure}[t]
    \fontsize{8}{10}\selectfont
         \centering
        \includegraphics[width = 0.48\textwidth]{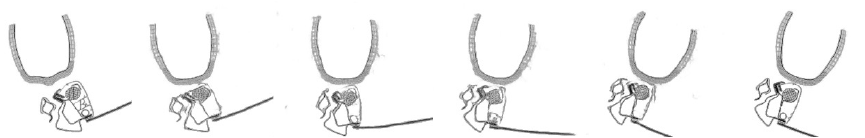}
\caption{Interpolation in latent space performs a rotation of the structure. The displayed image from left to right is sampled with the variable $\alpha \in [0, 0.2, 0.4, 0.6, 0.8, 1]$. }
\label{fig:figure_interpolation_images}
\end{figure}

\subsection{Interpolation in Latent Space}
After training, SA-ALAE can generate novel design alternatives based on a given blueprint. To test the potential of SA-ALAE in guiding the generation process and influencing the generated outputs, we identify two source images in which a similar object is rotated in the second image compared to the first. We show that various linear combinations of the latent variables used as input for the generator $G$ of the trained SA-ALAE$_{sub}$ model produce novel structures with various degrees of rotation, as shown in Fig.~\ref{fig:figure_interpolation_images}. The linear interpolation~\cite{b5, b9} between the latent variables encoded from two source images is obtained by the formula:
 \begin{equation}
 \label{interpolationFormulate}
\vec{\omega}_{\text{interpolated}} = (1-\alpha) \vec{\omega}_1 + \alpha \vec{\omega}_2,
\end{equation}
\noindent
where $\alpha \in [0,1]$ controls the interpolation and $\omega_1$, $\omega_2$ are the encoded latent variables. 


\section{Conclusion}
DGMs are known to synthesize photorealistic images. However, research in this area has been focusing on generating images of the natural world, such as faces and landscapes. As a result, there has been insufficient research on the effectiveness of DGMs in supporting industrial development processes. To address this challenge, our work proposes SA-ALAE by combining the ALAE's flexible framework, the attention mechanism's effect in large features, and modern approaches to stabilize adversarial training. In contrast to the previous work with GANs in synthesizing simple geometric shapes such as 2D curves, our research focuses on delivering an efficient solution for engineering processes. In addition to generating high-quality design images of complex structures, SA-ALAE allows users to extract latent variables from a given structural design and modify the given design by editing the latent variables. Despite the advances, several issues hinder an efficient application of SA-ALAE in Generative Engineering Design: (1) the current capability of SA-ALAE is insufficient for exploring designs for multiple vehicle components with a single model; (2) structural manipulation by modifying the latent space necessitates significant effort to analyze the learned latent variables and the structural attributes they represent.


\end{document}